\documentclass[10pt,twocolumn,letterpaper]{article}

 \usepackage{cvpr}              
\definecolor{cvprblue}{rgb}{0.21,0.49,0.74}
\usepackage[pagebackref,breaklinks,colorlinks,allcolors=cvprblue]{hyperref}

\def\paperID{*****} 
\def\confName{CVPR}
\def\confYear{2026}

\title{From Affect to Complex Behavior: Advancing Multimodal Human-Centered AI at the 10th ABAW Workshop \& Competition}

\author{Dimitrios Kollias \\
Queen Mary University of London (QMUL), UK\\
{\tt\small d.kollias@qmul.ac.uk}
\and
Panagiotis Tzirakis \\
Hume AI, USA \\
{\tt\small panagiotis@hume.ai}
\and
Alan Cowen\\
Google Deepmind, USA\\
\and
Stefanos Zafeiriou\\
Imperial College London, UK \\
\and
Irene Kotsia\\
Cogitat, UK \\
\and
Eric Granger\\
LIVIA, ILLS, ETS Montreal, Canada\\
{\tt\small eric.granger@etsmtl.ca}
\and
Marco Pedersoli\\
LIVIA, ILLS, ETS Montreal, Canada\\
\and
Simon Bacon\\
Concordia University, Canada\\
\and
Jens Madsen\\
Hume AI, USA \\
\and
Soufiane Belharbi \\
LIVIA, ETS Montreal, Canada\\
{\tt\small soufiane.belharbi@gmail.com}
\and
Muhammad Haseeb Aslam \\
LIVIA, ETS Montreal, Canada\\
\and
Chunchang Shao \\
QMUL, UK\\
\and
Guanyu Hu \\
QMUL, UK\\
Xi'an Jiaotong University, China 
}

\begin{document}
\maketitle
\begin{abstract}
The 10th Affective \& Behavior Analysis in-the-Wild (ABAW) Workshop and Competition, held at CVPR 2026, continues to advance research on modelling, analysis, understanding of human affect and behavior in real-world, unconstrained environments. The workshop maintains its dual structure, comprising both a competition and a paper track. The ABAW Competition introduces a diverse set of challenges targeting key aspects of affective and behavioral understanding, including continuous affect (valence-arousal) estimation, discrete affect (expression and action unit) recognition, as well as more complex behavior analysis tasks, such as emotional mimicry intensity estimation, ambivalence/hesitancy recognition and fine-grained violence detection. These challenges are built upon large-scale in-the-wild datasets, providing comprehensive benchmarks for state-of-the-art approaches. In parallel, the paper track presents a wide range of contributions spanning pose, motion \& behavior estimation, affect modelling \& multimodal learning, benchmarks, datasets \& evaluation protocols, fairness, robustness \& deployment. Overall, the 10th ABAW Workshop and Competition continues to serve as a key platform for benchmarking, collaboration and innovation, shaping the development of next-generation multimodal, human-centered AI systems.
\end{abstract}    
\section{Introduction}
\label{sec:intro}

Human affect and human behavior are central to how people communicate, make decisions, interact socially, and respond to their environments. Affect typically refers to internal emotional and expressive states, commonly studied through tasks such as valence-arousal estimation, facial expression recognition, micro-expression analysis, action unit detection, and physiological affect analysis. Human behavior, more broadly, includes observable actions and social signals such as body movement, gesture, pose, speech, vocal characteristics, attention, social interaction, mimicry, hesitation, and violence. Importantly, these phenomena are inherently multimodal: people do not communicate emotion and intent through a single channel, but through the dynamic combination of facial expressions, head movements, gaze, body gestures, voice, speech content, and sometimes physiological cues. As a result, modern AI systems increasingly aim to model affect and behavior through multimodal learning, where complementary signals from vision, audio, language, and other modalities are jointly exploited to achieve a richer and more reliable understanding of humans in real-world settings.

The importance of affect and behavior analysis is therefore both scientific and practical. Scientifically, it helps us better understand how emotions, intentions, and complex behaviors are expressed and perceived across different contexts. Practically, it underpins the development of multimodal human-centered AI systems that are socially aware, context-sensitive, and robust in unconstrained environments. Such capabilities are essential for applications in healthcare, education, human-computer interaction, robotics, media analysis, automotive systems, security, and digital humans. At the same time, multimodal human analysis introduces important challenges, including temporal dynamics, cross-modal fusion, missing or noisy modalities, demographic bias, privacy concerns, trustworthiness, and the need for fair and generalizable models. These challenges have driven the field beyond narrow unimodal recognition tasks toward richer multimodal formulations of affect and complex behavior in-the-wild.

Within this landscape, the ABAW workshop series has emerged as a major forum for advancing research on multimodal affect and behavior understanding in-the-wild. Over the years, the ABAW workshops have consistently brought together researchers working on analysis, modeling, understanding of human affect and behavior, while fostering interdisciplinary exchange across computer vision, machine learning, psychology, robotics, healthcare, ethics, human-machine interaction. As the field has evolved, so too has ABAW: from an initial emphasis on core affective computing tasks toward a broader and more ambitious agenda that includes richer behavioral phenomena, socially grounded human analysis, responsible multimodal AI.

The current event, the 10th ABAW Workshop and Competition, is held in conjunction with CVPR 2026 and reflects the trajectory of the field itself: from studying affect as a set of isolated labels to understanding more complex, temporally evolving, socially meaningful, and application-relevant aspects of human behavior. The workshop continues to serve not only as a benchmark-driven venue, but also as a platform for discussing how multimodal AI can support more robust, equitable, and human-centered technologies.

The accepted papers of the 10th ABAW Workshop further highlight the thematic diversity of the event and the broader trajectory of the field. They span pose, motion \& behavior estimation, affect modelling \& multimodal learning, benchmarks, datasets \& evaluation protocols, fairness, robustness \& deployment. Collectively, these contributions reinforce the workshop’s central message: progress in multimodal human-centered AI increasingly depends on moving beyond isolated affect recognition toward richer, temporally grounded, socially meaningful, and robust understanding of human behavior.

The ABAW series also stands out because of its associated competitions, which have played a central role in shaping benchmarks and driving progress in the community. Across its previous editions, the competitions have repeatedly focused on the core affective tasks of valence-arousal estimation, expression recognition, and action unit detection, while gradually expanding toward more challenging and behavior-oriented problems. Earlier editions introduced tasks such as multi-task learning, learning from synthetic data, emotional reaction intensity estimation, compound expression recognition, while more recent editions further broadened the scope to include emotional mimicry intensity estimation, ambivalence/hesitancy recognition, and fine-grained violence detection. This progression clearly illustrates how ABAW has evolved from foundational affect analysis toward richer tasks capturing nuanced, socially significant, and temporally structured human behaviors. In its 10th edition, ABAW continues this evolution through six challenges: Valence-Arousal Estimation, Expression Recognition, Action Unit Detection, Fine-Grained Violence Detection, Emotional Mimicry Intensity Estimation, and Ambivalence/Hesitancy Recognition. Taken together, these tasks span both well-established affective representations and more complex forms of behavior that require temporal, multimodal, and socially aware modeling.

In this paper, we present the 10th ABAW Workshop, summarize the ABAW Competition and position this workshop edition within the broader development of affective and behavior analysis in-the-wild. More broadly, this edition marks an important milestone for the ABAW series, reflecting a field that is increasingly multimodal, behavior-centric, and human-centered in both its scientific ambitions and its real-world impact.

\section{Workshop Overview}
The 10th ABAW Workshop brings together a diverse set of contributions spanning \textit{Pose, Motion and Behavior Estimation}, \textit{Affect Modelling and Multimodal Learning}, \textit{Benchmarks, Datasets and Evaluation Protocols}, and \textit{Fairness, Robustness and Deployment}. Below, we provide a brief overview of the accepted papers.

\subsection{Pose, Motion and Behavior Estimation}
\textit{"MuPPet: Multi-person 2D-to-3D Pose Lifting"} introduces a framework that explicitly models inter-person correlations, enabling more accurate and socially aware 3D pose estimation in multi-person scenarios. \textit{"VSDPose: Voxel-based Self-distillation for Multi-view 3D Human Pose Estimation"} leverages self-distillation within a voxel-based representation to improve multi-view 3D pose estimation. Complementing these, \textit{"Pose2Lang3D: Distilling 3D Reasoning from 2D Skeletons via Language Supervision"} learns language-aligned pose representations, enabling semantic reasoning from 2D inputs while benefiting from 3D supervision during training. In the context of action understanding, \textit{"SBF: Augmenting Skeleton for Effective Video-based Human Action Recognition"} enriches skeleton-based pipelines with depth, body contour and motion cues, significantly improving recognition performance. \textit{"VGGT-HPE: Reframing Head Pose Estimation as Relative Pose Prediction"} reformulates head pose estimation as a relative prediction problem, demonstrating strong performance using a geometry-based foundation model trained on synthetic data.

\subsection{Affect Modelling and Multimodal Learning}

\textit{"LaScA: Language-Conditioned Scalable Modelling of Affective Dynamics"} integrates language-derived semantic representations with structured features to improve affect dynamics' modelling in lightweight manner. \textit{"Beyond the Mean: Modelling Annotation Distributions in Continuous Affect Prediction"} proposes a distribution-aware framework that captures annotator variability and uncertainty, moving beyond point estimates. In physiological sensing, \textit{"Trust What You Fuse: Reliability-Aware Cross-Attention for Multimodal Physiological Stress Assessment in the Wild"} introduces a reliability-aware fusion mechanism that dynamically weights modalities under noisy real-world conditions. \textit{"Dimensional Distribution Emotion State: Leveraging Valence and Arousal as a Common Embedding Space for Visual Emotion Analysis"} proposes a novel representation for modelling emotional responses, enabling richer and more flexible emotion embeddings across datasets.

\subsection{Benchmarks, Datasets \& Evaluation Protocols}

\textit{"SenBen: Sensitive Scene Graphs for Explainable Content Moderation"} introduces a large-scale benchmark with grounded scene graph annotations for sensitive content, enabling more interpretable and explainable content moderation systems. \textit{"MTLLFM: Multimodal-Temporal Laughter Localization: UR-FUNNY-Temporal and SMILE-Temporal Benchmarks with an Adaptive Multimodal Fusion Model"} provides temporally annotated datasets for laughter localization and proposes a multimodal model for fine-grained temporal grounding. \textit{"From Frames to Events: Rethinking Evaluation in Human-Centric Video Anomaly Detection"} highlights the limitations of frame-level metrics and introduces event-centric evaluation protocols for anomaly detection. \textit{"Beyond the Fold: Quantifying Split-Level Noise and the Case for Leave-One-Dataset-Out AU Evaluation"} systematically analyzes the variability introduced by standard cross-validation protocols and advocates for more stable evaluation through leave-one-dataset-out settings.

\subsection{Fairness, Robustness and Deployment}
 \textit{"Deconfounding demographic bias estimation in facial expression recognition"} demonstrates that visual confounders such as head pose and illumination can significantly distort fairness assessments, and proposes a standardized evaluation protocol.  \textit{"Topology-Guided Test-Time Adaptation via Persistent Homology: From Affective Behavior Analysis to Autonomous Driving"} introduces a principled, topology-aware framework for determining when test-time adaptation should be applied, improving robustness under domain shift without degrading performance.

\section{Competition Overview}
Detailed information about the competition is available at the following webpage, which also includes the leaderboards for each challenge and links to  participating teams' GitHub repositories, where the source code of their methods is provided:
\url{https://affective-behavior-analysis-in-the-wild.github.io/10th}

\subsection{Valence-Arousal (VA) Estimation Challenge}

\paragraph{Description}
This challenge focused on per-frame  prediction of two continuous affect dimensions, valence (ranging from positivity to negativity) and arousal (ranging from  from passive to active activity levels). 

\begin{figure}[h]
\centering
\includegraphics[height=6cm]{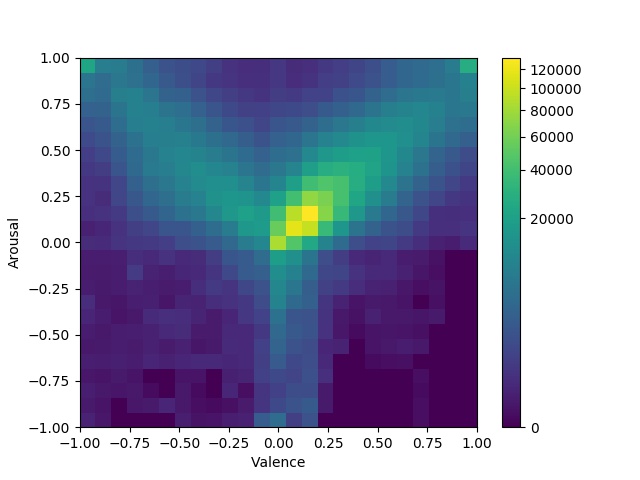}
\caption{VA Estimation Challenge: 2D VA Histogram}
\label{va_annot}
\end{figure}

\paragraph{Database}
This Challenge’s dataset comprised 594 videos, an expansion of the Aff-Wild2 database \cite{kollias2019deep,kollias2025advancements,kollias2025emotions}, annotated in terms of valence and arousal. These annotations span 2,993,081 frames from 584 subjects, with 16 videos featuring two subjects, both of whom are annotated. The annotations were conducted by four experts using the methodology described in \cite{cowie2000feeltrace}. Valence and arousal values are continuous and range from -1 to 1. The 2D Valence-Arousal histogram of the dataset is presented in Figure \ref{va_annot}.

The dataset was divided into training, validation and testing sets, ensuring subject independence, meaning each subject appears in only one set. The splits are:
356 videos in the training set; 76 videos in the Validation set; 162 videos in the testing set.
The training and validation sets, along with their annotations, were provided to participants. The test set was unlabeled, and participants should have submitted their predictions to an evaluation server. Each team was allowed to have up to five submissions.

\paragraph{Data Pre-processing}
All videos in Aff-Wild2 were first decomposed into individual frames; these were  processed using Behavior4All toolkit's \cite{kollias2024behaviour4all} detector to extract face bounding boxes and five facial landmarks (two eyes, nose, and two mouth corners) for each frame. The detected bounding boxes were used to crop the faces, and these cropped images were provided to the participants.

To ensure spatial alignment, we also used Behavior4All toolkit for creating cropped and aligned images, which were also made available to participants. 

These processed images are utilized in our baseline experiments. These images are resized to $112 \times 112 \times 3$ pixels, and their intensity values are normalized to the range [-1, 1].

\paragraph{Rules}
Only uni-task solutions were permitted for this challenge. Teams could utilize any publicly or privately available pre-trained model, provided it had not been trained on the Aff-Wild2 dataset used in this challenge. Pre-trained models could be based on any task, such as valence-arousal estimation, expression recognition, action unit detection, or face recognition. However, during model refinement and methodology development, teams should have strictly used only valence-arousal annotations and should not have incorporated expression or action unit labels. Acceptable data augmentation strategies included using valence-arousal annotations from other datasets, synthetic or generated data, affine transformations, and other augmentation techniques for enhancing training.

\paragraph{Evaluation Metric}
The performance metric is the average Concordance Correlation Coefficient (CCC) for valence and arousal:

\begin{equation} 
\mathcal{P}_{VA} = \frac{CCC_a + CCC_v}{2}
\end{equation}

The CCC measures the agreement between two time series (e.g., annotations and predictions) while penalizing deviations in both correlation and mean squared differences. CCC values range from -1 to 1. High values are desired.
The CCC formula is defined as follows:

\begin{equation} \label{ccc}
CCC  = \frac{2s_x  s_y \rho_{xy}}{s_x^2 + s_y^2 + (\bar{x} - \bar{y})^2},
\end{equation}

\noindent
where $\rho_{xy}$ is the Pearson's Correlation Coefficient, $s_x$ and $s_y$ represent the variances of valence/arousal annotations and predicted values, respectively, and $s_{xy}$ is the corresponding covariance value.

\paragraph{Baseline}  
The baseline model is based on ResNet-50, pre-trained on ImageNet. The baseline model achieved an average CCC of 0.22 on the validation set.

\paragraph{Top-3 Teams}  
The winning team was \textit{RAS} \cite{ryumina2026team1}. They proposed a multimodal framework that combines facial, behavioral, audio cues. Facial features were extracted with GRADA encoder and modeled temporally with Transformer; behavioral features were obtained using Qwen3-VL and modeled with Mamba; audio features were extracted with WavLM. The modalities were fused using cross-modal mixture-of-experts or a reliability-aware fusion scheme.

The runner-up was team \textit{EmoDX} \cite{lee2026stage}. They proposed SAGE, a multimodal framework that explicitly models modality reliability. Visual features were extracted with ResNet-50; audio features with WavLM, followed by TCN to capture short-term dynamics. A stage-adaptive module dynamically reweighted A/V contributions via reliability-guided fusion and Transformer-based temporal refinement. 

In the third place was team \textit{IMLAB} \cite{jung2026distance}. They proposed a multimodal framework that integrates visual and audio features with semantic guidance. Visual features were extracted using CLIP encoder, enhanced via distance-aware soft prompt mechanism that aligns VA values with semantic region prototypes; audio features were obtained with AST. Both modalities were temporally modeled using GRUs and fused via hierarchical cross-modal attention \& gated fusion. 

\paragraph{Results}
Table \ref{va} compares the performance on the test set of the top-4 methods and the baseline, in terms of total CCC.

\begin{table}[h]
\caption{VA Estimation Challenge
 Results} 
\label{va}
\centering
\scalebox{0.9}{
\begin{tabular}{ |c||c| }
 \hline
 Teams  & 
Total Score  
\\ 
  \hline
 \hline

RAS \cite{ryumina2026team1} & 0.62  \\ 
\hline

EmoDX \cite{lee2026stage} & 0.58  \\ 
\hline

IMLAB \cite{jung2026distance} & 0.53  \\ 
\hline

HSEmotion \cite{savchenko2026hsemotion} & 0.52   \\ 
\hline

baseline  
&  0.20
 \\
\hline

\end{tabular}
}
\end{table}

\subsection{Expression (EXPR) Recognition Challenge}

\paragraph{Description}
The objective of this challenge is to classify expressions in each frame into 8 mutually exclusive categories: 6 basic expressions (anger, disgust, fear, happiness, sadness, surprise), neutral state, and "other" category for non-basic expressions.

\paragraph{Database}
This challenge utilized 548 videos from the Aff-Wild2 database. Seven videos contain two subjects, both of whom were annotated. In total, the dataset included 2,624,160 frames from 437 subjects. Annotations were conducted frame-by-frame by seven experts.
Table \ref{expr_distr} presents the distribution of expression annotations.

\begin{table}[!h]
\caption{EXPR Recognition Challenge: Expressions distribution}
\label{expr_distr}
\centering
\scalebox{0.9}{
\begin{tabular}{ |c||c| }
\hline
 Expressions & No of Images \\
\hline
\hline
Neutral & 468,069  \\
 \hline
Anger & 36,627  \\
 \hline
Disgust & 24,412 \\
 \hline
Fear &  19,830 \\
 \hline
Happiness & 245,031  \\
 \hline
Sadness & 130,128  \\
 \hline
Surprise & 68,077  \\
 \hline
 Other & 512,262 \\
 \hline
\end{tabular}
}
\end{table}

The dataset was divided into training, validation, and testing sets, ensuring subject independence, meaning each subject appears in only one set. The splits were:
248 videos in the training set; 70 videos in the Validation set; 230 videos in the testing set. For sharing these, same procedure was followed as in the VA Estimation Challenge. Each team was allowed to have up to five submissions.

\paragraph{Data Pre-processing}
The same data pre-processing was conducted as described for the VA Estimation Challenge.

\paragraph{Rules}
Similar rules to the VA Estimation Challenge apply here as well.

\paragraph{Evaluation Metric}
The performance metric is the macro $F1$ score across all 8 classes:

\begin{equation} 
\mathcal{P}_{EXPR} = \frac{\sum_{expr} F_1^{expr}}{8}
\end{equation}

The $F1$ score balances precision (proportion of correctly identified positive samples) and recall (proportion of actual positive samples correctly identified). The $F1$ score is:

\begin{equation} \label{f1}
F_1 = \frac{2 \times precision \times recall}{precision + recall}
\end{equation}

\paragraph{Baseline}
The baseline model is based on a pre-trained on VGGFace VGG16 with non-trainable conv layers (3 fc layers are trainable).  MixAugment \cite{psaroudakis2022mixaugment} was used for data augmentation. Its peformance on the validation set was 0.25.

\paragraph{Top-3 Teams}
The winning team was \textit{EagleonPamir1} \cite{sun2026two}. They proposed a two-stage dual-modality framework. In Stage I, a DINOv2 visual encoder was fine-tuned using padding-aware augmentation \& a training-only MoE head.
In Stage II, multi-scale facial features were extracted \& combined with frame-aligned audio features from Wav2Vec 2.0.
A  gated fusion module integrated the 2 modalities; inference-time temporal smoothing was also applied.

The runner-up was team \textit{HSEmotion} \cite{savchenko2026hsemotion}. They proposed a confidence-gated calibration framework. Facial embeddings were extracted using pretrained EfficientNet-based emotion models. If model’s confidence exceeded a threshold, its prediction was directly used; otherwise, embeddings were passed to lightweight MLP trained on Aff-Wild2.
Logit adjustment was applied for calibratation. 

In the third place was team \textit{USTC-IAT-United} \cite{yu2026solution}. They proposed a  multimodal framework tackling missing modalities and class imbalance. Visual and audio features were extracted using BEiT-large and WavLM-large.
A dual-branch Transformer with bidirectional cross-attention and learnable gating dynamically fused the modalities.
To handle occlusions, modality dropout and cross-attention mechanism enabled degradation to audio-only predictions.

\paragraph{Results}
Table \ref{expr} compares the performance on the test set of the top-4 methods and the baseline, w.r.t. macro F1 score.

\begin{table}[h]
\caption{EXPR Recognition Challenge
 Results} 
\label{expr}
\centering
\scalebox{0.9}{
\begin{tabular}{ |c||c| }
 \hline
 Teams  & 
Total Score 
\\ 
  \hline
 \hline

EagleonPamir1 \cite{sun2026two} & 0.391  \\ 
\hline

HSEmotion \cite{savchenko2026hsemotion} & 0.386  \\ 
\hline

USTC-IAT-United \cite{yu2026solution} & 0.36  \\ 
\hline

IMLAB \cite{byeon2026multimodal} & 0.32  \\ 
\hline

baseline  
&  0.225
  
 \\
\hline

\end{tabular}
}
\end{table}

\subsection{Action Unit (AU) Detection Challenge}

\paragraph{Description}
The objective of this challenge is to detect the presence of 12 AUs in each video frame. AUs correspond to specific facial muscle movements, such as inner and outer brow raisers, cheek raisers, lip tighteners, and jaw drops.  The specific AUs are: AU1, AU2, AU4, AU6, AU7, AU10, AU12, AU15, AU23, AU24, AU25 and AU26. 

\paragraph{Database}
This challenge utilized 542 videos from the Aff-Wild2 database, annotated for 12 AUs, corresponding to key facial muscle movements such as brow raisers, cheek raisers, lip tighteners, and jaw drops. In total, the dataset included 2,627,632 frames from 438 subjects, with annotations generated through a semi-automatic process combining manual and automatic techniques.
Table \ref{au_distr} outlines the distribution of AU annotations.

\begin{table}[h]
    \centering
        \caption{AU Detection Challenge: Distribution of AU Annotations}
    \label{au_distr}
    \scalebox{0.8}{
\begin{tabular}{|c|c|c|}
\hline
  Action Unit \# & Action   &\begin{tabular}{@{}c@{}} Total Number \\  of Activated AUs \end{tabular} \\   \hline    
    \hline    
   AU 1 & inner brow raiser   & 301,102 \\   \hline    
   AU 2 & outer brow raiser  & 139,936 \\   \hline   
   AU 4 & brow lowerer   & 386,689  \\  \hline    
   AU 6 & cheek raiser  & 619,775 \\  \hline    
   AU 7 & lid tightener  & 964,312 \\  \hline    
   AU 10 & upper lip raiser  & 854,519 \\  \hline    
   AU 12 & lip corner puller  & 602,835 \\  \hline    
   AU 15 & lip corner depressor  & 63,230 \\  \hline   
  AU 23 & lip tightener & 78,649 \\  \hline    
   AU 24 & lip pressor & 61,500 \\  \hline    
   AU 25 & lips part  & 1,596,055 \\  \hline     
   AU 26 & jaw drop  & 206,535 \\  \hline     
\end{tabular}
}
\end{table}

The dataset was divided into training, validation, and testing sets, ensuring subject independence, meaning each subject appears in only one set. The splits were:
295 videos in the training set; 105 videos in the Validation set; 142 videos in the testing set.
For sharing these, same procedure was followed as in the VA Estimation Challenge. Each team was allowed to have up to five submissions.

\paragraph{Data Pre-processing}
The same data pre-processing was conducted as described for the VA Estimation Challenge.

\paragraph{Rules}
Similar rules to the VA Estimation Challenge apply here as well.
Only uni-task solutions were permitted for this challenge. Teams could utilize any publicly or privately available pre-trained model, provided it had not been trained on the Aff-Wild2 dataset used in this challenge. During model refinement and methodology development, teams should have strictly used only AU annotations and  not incorporate VA or EXPR labels.

\paragraph{Evaluation Metric}
The performance measure is the average F1 Score across all 12 AUs:

\begin{equation} 
\mathcal{P}_{AU} = \frac{\sum_{au} F_1^{au}}{12}
\end{equation}

\paragraph{Baseline}
The baseline follows the same VGG16-based architecture used in the EXPR Recognition Challenge.  Its average F1 score on the validation set was 0.39.

\paragraph{Top-2 Teams} Here we present the top-2 teams that surpassed the performance of the baseline model. 
The winning team was \textit{USTC-IAT-United} \cite{yu2026hierarchical}. They proposed a multimodal framework that combined pretrained DINOv2 for vision and WavLM for audio with hierarchical alignment and long-range temporal modeling. A Hierarchical Granularity Alignment module aligned local facial regions with global facial context using cross-attention. Temporal dynamics were modeled using an audio-guided state space model that integrated audio cues to guide visual updates.

The runner-up was team \textit{HSEmotion} \cite{savchenko2026hsemotion}. They proposed an efficient framework built on strong pretrained representations with task-specific heads. Facial embeddings from pretrained EmotiEffLib were fed into a lightweight MLP. The approach focused on calibration: (i) per-AU class weights were applied, (ii) optimal decision thresholds were tuned individually for each AU, and (iii) temporal smoothing was applied. Both embedding-based and logit-based representations were combined.

\paragraph{Results}
Table \ref{au} compares the performance on the test set of of the top-2 methods (that surpassed the baseline performance) and the baseline, in terms of average F1 score.

\begin{table}[h]
\caption{AU Detection Challenge
 Results} 
\label{au}
\centering
\scalebox{0.85}{
\begin{tabular}{ |c||c| }
 \hline
 Teams & Total Score  
\\ 
  \hline
 \hline

USTC-IAT-United \cite{yu2026hierarchical} & 0.51   \\ 
\hline

HSEmotion \cite{savchenko2026hsemotion} & 0.49 \\ 
\hline

baseline  
&  0.365

 \\
\hline

\end{tabular}
}
\end{table}

\subsection{Fine-Grained Violence Detection (VD) Challenge}

\paragraph{Description}
The objective of this Challenge is to distinguish violent from non-violent frames in videos of diverse and challenging scenarios.

\paragraph{Database}  
This challenge utilizes a subset of DVD database \cite{kollias2025dvd,kollias2025emotions}. DVD is an A/V, large-scale, in-the-wild, frame-level annotated violence detection dataset with over 500 videos (2.8M frames). It captures diverse environments, varying lighting conditions, multiple camera sources, complex social interactions and rich metadata, reflecting the complexity of real-world violent events.
The subset utilized in this challenge  consisted of 172 videos, each annotated in a per-frame basis for the presence or absence of violence. In total annotations are provided for 1,389,976 frames. Table \ref{dvd} outlines their distributions.

\begin{table}[h]
    \centering
        \caption{Fine-Grained VD Challenge: Distribution of Annotations}
    \label{dvd}
    \scalebox{0.9}{
\begin{tabular}{|c||c|}
\hline
  Labels  &  Total \# \\  \hline    
    \hline    
   Violent & 598,584 \\   \hline    
   non-Violent & 761,970 \\   \hline   
\end{tabular}
}
\end{table}

The dataset was divided into training, validation and testing sets, ensuring subject independence, meaning each subject appears in only one set. The splits were:
103 videos in the training set; 17 videos in the Validation set; 52 videos in the testing set.
For sharing these, same procedure was followed as in the VA Estimation Challenge. Each team was allowed to have up to five submissions.

\paragraph{Evaluation Metric}  
The performance metric is the macro F1 Score across the two categories.

\paragraph{Baseline}  
The baseline for this challenge is a ResNet-50 (pre-trained on ImageNet) plus one-layer bidirectional LSTM. Its macro F1 Score on the validation set is 0.64.

\paragraph{Winner}  
Here we present the winning team that surpassed the performance of the baseline. The winning team for this challenge was \textit{HSEmotion}. They propose a multimodal framework comprising a visual and skeleton stream. The visual stream employs a ConvNeXt-T backbone followed by a dilated TCN; the skeleton stream encodes keypoints, motion (velocities) and interaction features extracted via MediaPipe. The two modalities are projected into a shared space and fused via bidirectional cross-attention; the resulting representation is decoded via 2-layer BiLSTM.

\paragraph{Results}  
Table \ref{vd} compares the performance on the test set of the winning method (that surpassed the baseline performance) and the baseline, in terms of macro F1 Score.

\begin{table}[h]
\caption{Fine-Grained VD Challenge
 Results} 
\label{vd}
\centering
\scalebox{0.9}{
\begin{tabular}{ |c||c| }
 \hline
 Teams  & 
 Total Score  
\\ 
  \hline
 \hline

HSEmotion & 0.587  \\
\hline

baseline  
&  0.504

 \\
\hline

\end{tabular}
}
\end{table}

\subsection{Emotional Mimicry Intensity (EMI) Estimation Challenge}

\paragraph{Description}
This challenge explores emotional mimicry by predicting the intensity of the following six emotional dimensions: Admiration, Amusement, Determination, Empathic Pain, Excitement and Joy.

\paragraph{Database}  
This challenge utilizes HUME-Vidmimic2 dataset,  consisting of 15,000+ videos totaling 30+ hours from 557 participants. The dataset was collected in naturalistic settings, where participants used their webcams to mimic facial and vocal expressions from seed videos and rate their mimicry on 0-100 scale (a normalized score from 0 to 1 is provided). The dataset is partitioned as seen in Table \ref{emi_partitions}. The dataset ensures speaker independence, i.e., no participant appears in more than one partition.

\begin{table}[h!]
\centering
\caption{EMI Estimation
Challenge statistics}
\scalebox{0.9}{
\begin{tabular}{|l|r|r|}
\hline
Partition & Duration & \# Samples \\ 
 & (HH:MM:SS) & \\ \hline \hline
Train & 15:07:03 & 8072 \\ \hline
Validation & 9:12:02 & 4588 \\ \hline
Test & 9:04:05 & 4586 \\ \hline
\end{tabular}
}
\label{emi_partitions}
\end{table}

\paragraph{Data Pre-processing}
Participants are provided with facial detections from MTCNN, processed at a frequency of 6 frames/second. Additionally, participants are provided with pre-extracted features from: i) Faces: Processed using ViT; ii) Audio Signals: Processed using Wav2Vec 2.0.

\paragraph{Evaluation Metric}  
The performance metric is the average Pearson’s Correlation Coefficient ($\rho$) across 6 emotions:

\begin{equation} \label{ppp}
\mathcal{P}_{EMI} = \frac{\sum_{i=1}^{6} \rho^{i}}{6}
\end{equation}


\paragraph{Baseline}  
We set initial baseline results with two distinct sets of features: i) Vision-based features extracted from a pre-trained ViT; ii) Audio-based features extracted using Wav2Vec2, followed by a linear layer. Each feature set was processed separately using a three-layer GRU. The vision-only, audio-only and multimodal models achieved on the validation set 0.09, 0.24 and 0.255, respectively.

\paragraph{Top-3 Teams}  
The winning team was \textit{USTC-IAT-United} \cite{zhu2026anchoring}. They propose TAEMI, a multimodal framework that integrates visual, acoustic and textual modalities. It employs a Text-Anchored Dual Cross-Attention mechanism, where textual features guide the alignment and filtering of noisy visual and audio signals. Each modality is encoded and projected into shared space and modeled with self-attention before cross-modal fusion.  The method  introduces learnable missing-modality tokens and modality dropout.

The runner-up was team \textit{CASIA26} \cite{huang2026multimodal}. They proposed a multimodal framework  based on pretrained visual, audio and text features. It adopted a simple feature-level concatenation to preserve modality-specific information. The model was trained using a multi-objective loss combining MSE, Pearson correlation and auxiliary supervision. Additionally, a VAD-inspired latent prior was introduced to enhance acoustic representations.

In the third place was team \textit{MimicMetric}. They proposed a two-stage multimodal framework that combined text, audio, visual and optional motion features. In Stage I, modality-specific encoders (DINOv2 for vision, wav2vec2 for audio, pretrained text embeddings) were trained independently. In Stage II, the learned embeddings were concatenated. Temporal dynamics were captured via attention pooling in audio, vision and motion branches.

\paragraph{Results}  
Table \ref{emi} compares the performance on the test set of the top-3 methods and the baseline, w.r.t. average $\rho$.

\begin{table}[h]
\caption{EMI Estimation Challenge
 Results} 
\label{emi}
\centering
\scalebox{0.9}{
\begin{tabular}{ |c||c| }
 \hline
Teams  & Total Score  
\\ 
  \hline
 \hline

USTC-IAT-United \cite{zhu2026anchoring} & 0.708  \\ 
\hline

CASIA26 \cite{huang2026multimodal} & 0.674   \\ 
\hline

MimicMetric & 0.57  \\ 
\hline

baseline (A / V / AV) 
&  0.27 / 0.13 / 0.29  
 \\
\hline

\end{tabular}
}
\end{table}

\subsection{Ambivalence/Hesitancy (AH) Video Recognition Challenge}

\paragraph{Description}
This challenge aims to predict Ambivalence/Hesitancy (A/H) at the video level, formulating it as a binary classification task where each video is labeled as either containing A/H (1) or not (0).

\paragraph{Database}
This Challenge’s dataset was a new, fully annotated at video- and frame-level version of  BAH dataset \cite{bahdb}, collected for multimodal recognition of A/H in videos. It contained 1,427 videos with a total duration of 10.60 hours, captured from 300 participants across Canada, answering a predefined set of questions to elicit A/H. It was intended to mirror real-world online personalized behaviour change interventions. BAH is fully annotated by experts to provide timestamps that indicate where A/H occurs, frame- and video-level annotations with A/H cues. Speech-to-text transcripts, their timestamps, cropped and aligned faces, participants' metadata are also provided. Each dataset participant may have up to 7 videos. Table \ref{bah_db_distr} presents the distributions.

\begin{table}[h]
\caption{AH Video Recognition Challenge: Distributions }
\centering
\label{bah_db_distr}
\scalebox{0.8}{
\begin{tabular}{lcccc}
\hline
\textbf{Data subsets} & \textbf{Train} & \textbf{Validation} & \textbf{Test}  \\
\hline
Number of participants & 195 & 30 & 75  \\\hline
Number of participants with A/H & 144 & 27 & 75  \\\hline
Number of videos & 778 & 124 & 525  \\\hline
Number of videos with A/H & 385 & 75 & 318  \\\hline
Number of frames & 501,970 & 79,538 & 335,110  \\\hline
Number of frames with A/H & 76,515 & 13,984 & 65,756 \\\hline
Total duration (hour) & 5.80 & 0.92 & 3.87  \\\hline
Total duration with A/H (hour) & 0.87 & 0.16 & 0.75  \\
\hline
\end{tabular}
}
\end{table}

The dataset is divided participant-wise into: train (195 participants), public test (30 participants), private test (75 participants) sets. Videos of one participant belong to only one set. 
The training and public test sets and
their annotations, were provided to participants. The private test set was unlabeled, participants should  submit their predictions to the organizers. Each team was allowed to have up to five submissions, either all at once, or one  at a time. 

\paragraph{Evaluation Metric}
The performance metric is the macro F1 score (at video level) across both classes.

\paragraph{Baseline}  
The baseline model is a zero-shot setup with Video-LLaVA with a simple prompt and vision modality only. Its performance on the public test set is 0.2827.

\paragraph{Top-3 Teams}  
The winning team was \textit{VisPBF} \cite{pereira2026brother}. They proposed  BROTHER, an ensemble framework that combined multiple modality-specific models to capture complementary behavioral cues. It integrated visual, audio,  contextual features via separate expert networks. It introduced ensemble diversity constraints and consistency-based regularization across predictors. Temporal information was incorporated via sequence modeling modules. 

The runner-up was team \textit{Fennec} \cite{bekhouche2026conflict}. They proposed ConflictAwareAH, a multimodal framework  that explicitly modelled cross-modal inconsistencies. It used pretrained encoders (VideoMAE, HuBERT, RoBERTa-GoEmotions) to extract video, audio, text embeddings, which are aggregated via attention pooling. The key idea was to compute pairwise conflict features as absolute differences between modality embeddings, capturing disagreement signals. 

In the third place was team \textit{LEYA} \cite{ryumina2026team}. They proposed a multimodal framework that leverages visual, audio and textual cues. Modality-specific features are extracted using pretrained encoders and then refined via temporal modules (LSTM/attention). The model introduces a modality-aware fusion mechanism that adaptively weights each modality based on its reliability.  It also incorporates auxiliary supervision and regularization strategies.

\paragraph{Results}
Table \ref{bahh} compares the performance on private test of the top-5 methods and baseline, w.r.t. macro F1.

\begin{table}[h]
\caption{AH Video Recognition Challenge
 Results} 
\label{bahh}
\centering
\scalebox{0.9}{
\begin{tabular}{ |c||c| }
 \hline
 Teams  & 
Total Score  
\\ 
  \hline
 \hline

VisPBF \cite{pereira2026brother}           &  0.7266               \\ 
\hline

Fennec \cite{bekhouche2026conflict}          &  0.7151          \\ 
\hline 

LEYA   \cite{ryumina2026team}          &  0.7142             \\ 
\hline 

Lenovo PCIE \cite{tang2026nuanced}      &  0.6748              \\ 
\hline 

Time Visão \cite{souza2026solution}      &  0.5362           \\
\hline

Baseline         &  0.3428          \\ 
\hline

\end{tabular}
}
\end{table}

\section{Conclusion}

The 10th ABAW Workshop and Competition highlights the continued growth of the field from core affect recognition toward richer, multimodal, temporally grounded and socially meaningful analysis of human behavior in-the-wild. Through its six competition challenges and its diverse paper track, the workshop provides a valuable platform for benchmarking, collaboration and innovation, supporting the development of multimodal, robust, and human-centered AI systems. The results of this edition further demonstrate the importance of multimodal fusion, strong pretrained models, temporal reasoning and deployment-aware design in addressing real-world affective and behavioral analysis tasks.

{
    \small
    \bibliographystyle{ieeenat_fullname}
    \bibliography{main}
}


\end{document}